\documentclass[a4paper, 10pt, twocolumn, twoside]{article}

\usepackage{ISARC}
\usepackage{lscape}
\usepackage{hologo}
\usepackage{array}
\newcolumntype{L}[1]{>{\raggedright\arraybackslash}p{#1}}
\begin{document}
\linespread{0.5}

\title{Adaptive Human--Robot Collaboration for Masonry Construction Under Material and Assembly Uncertainty}

\author{Jutang Gao$^{1}$ and Arash Adel$^{1}$*}

\affiliation{
$^{1}$Princeton University, USA \\
*Corresponding Author
}

\email{
\href{mailto:jutang.gao@princeton.edu}{jutang.gao@princeton.edu},
\href{mailto:arash.adel@princeton.edu}{arash.adel@princeton.edu}
}

\maketitle
\pagestyle{plain}

\begin{abstract}
Human--robot collaboration in construction is often challenged by limited robot-to-human communication and the need to adapt to tolerance accumulation arising from material and assembly uncertainties. We present an adaptive human--robot collaborative workflow for masonry construction that addresses communication limitations and tolerance accumulation, demonstrated through a brickwork case study in which a robot places bricks while a human applies adhesive. This workflow is enabled by two complementary mechanisms: 1) an end-effector-mounted projector that provides spatially registered, just-in-time projection guidance for manual adhesive application, and 2) laser scanning for feedback-driven grasping and placement pose correction. Together, these mechanisms enable adjustment of human and robotic actions in response to material variability and accumulated assembly tolerances. Full-scale experiments across conventional running-bond and nonstandard configurations demonstrate that projection guidance improves adhesive application consistency and reduces application time, while laser-based correction maintains level courses and avoids collision-prone failures associated with open-loop execution. These results indicate that integrating spatial projection with feedback-driven adaptation, enabled by material and as-built sensing, can mitigate tolerance accumulation and improve precision and robustness in human--robot collaborative construction.
\end{abstract}


\section{Introduction}
\label{sec:introduction}

Robotic systems are increasingly viewed as a means to improve productivity, safety, and working conditions in construction, while enabling mass customization, improved building performance, and novel architecture~\cite{adel2024, liang_humanrobot_2021, Adel2022, Craney2020}. Yet the practical deployment of construction robotics remains limited, partly because robots must be integrated into human-centered workflows~\cite{davila_delgado_robotics_2019} and operate in unstructured, dynamic environments where material imperfections, as-built deviations, and environmental uncertainties challenge open-loop execution~\cite{chen2025, adel2024, Ruan2023}. Human--robot collaboration has therefore been proposed and demonstrated as a way to improve adaptability, flexibility, and robustness by combining human dexterity, situated judgment, and material handling capabilities with robotic precision and repeatability~\cite{kyjanek_implementation_2019, Adel2020, Yang2024}. Nevertheless, such collaboration remains constrained by limited and often non-intuitive methods for information exchange and task handoff~\cite{kyjanek_implementation_2019}, as well as limited robotic adaptability to material and as-built deviations~\cite{adel2024}. Addressing this gap requires workflows that provide spatially explicit robot-to-human guidance while enabling feedback-driven robotic correction based on measured deviations.

This paper uses adhesive-based brickwork as a case study of human--robot collaborative masonry construction, where a human applies adhesive and the robot places bricks. While open-loop robotic brickwork has been explored in prior research~\cite{Bonwetsch2015}, it is prone to tolerance accumulation due to variability in brick dimensions and placement inaccuracies. Unlike mortar-based masonry, where such tolerances can be compensated during placement, adhesive-based brickwork is far less tolerant due to thin bonding layers. This case study, therefore, highlights the necessity of feedback-driven, adaptive construction workflows. Accordingly, the paper's main contributions are:
\begin{itemize}[nosep]
\item We present a human--robot collaborative brickwork workflow that integrates projection-based robot-to-human guidance for adhesive application.
\item We introduce an adaptive brick pose correction loop that uses laser-based sensing to detect placement tolerances and enforce explicit geometric constraints in the presence of uncertainty in brick and adhesive placement. 
\item We evaluate the workflow through full-scale physical experiments, quantifying performance in terms of adhesive application accuracy, tolerance handling, and cycle times for human and robot actions.
\end{itemize}
Altogether, this work proposes a transferable workflow for adaptive human--robot collaborative masonry construction under material variability and tolerance accumulation, which are inherent to real-world construction.

\section{Related Work}
\label{sec:background}

This section contextualizes our work within three bodies of literature: human--robot collaboration in construction (Section~\ref{sec:HRC}), spatial augmented reality for information sharing (Section~\ref{sec:spatialAR}), and feedback-driven adaptive approaches for construction (Section~\ref{sec:adaptive}). 

\subsection{Human--Robot Collaboration in Construction}
\label{sec:HRC}

Human--robot collaboration involves humans and robots working together to leverage complementary capabilities~\cite{liang_humanrobot_2021}, combining robotic precision with human decision-making and adaptability. 
Prior research has emphasized the importance of legible and predictable robot motion for conveying intent~\cite{dragan_legibility_2013}, as well as communication interfaces that can enhance safety, productivity, and quality in collaborative processes~\cite{kyjanek_implementation_2019, zhang_projected_2022, brosque_human-robot_2020}. However, many existing implementations still rely primarily on one-way information flow from humans to robots through direct commands, programming, and supervision~\cite{liang_humanrobot_2021}. To address this gap, this paper investigates just-in-time robot-to-human communication through spatially registered projection onto the work-in-progress structure to guide human actions in collaborative masonry.

\subsection{Projection-Based Augmented Reality for Information Sharing}
\label{sec:spatialAR}

Prior work on projection-based spatial augmented reality for human--machine communication explores diverse information types, projection devices, and spatial configurations for effective information delivery~\cite{zhang_projected_2022, ahn_2d_2019, degani_automated_2019, tavares_collaborative_2019}. From this set of work, two key trade-offs emerge: 1) projection reach versus cost: high-power projectors provide extended reach and higher resolution at increased cost and energy demand, whereas low-power projectors are more cost-efficient but constrained in visibility range; 2) flexibility versus accuracy: fixed setups provide higher accuracy, whereas mobile setups provide extra flexibility but introduce localization challenges. This paper approaches this problem by mounting a low-cost, low-power projector on the robot end-effector and leveraging robotic positioning accuracy to enable flexible, spatially registered projection to aid human--robot collaborative masonry. 

\subsection{Feedback-Driven Adaptive Construction}
\label{sec:adaptive}

Adaptive construction allows robots to adjust actions based on measured physical tolerances rather than executing open-loop plans~\cite{adel2024}. Discrete assembly tasks in construction, such as brickwork, inherently involve tolerances in materials and processes, leading to accumulative errors if uncorrected~\cite{Bonwetsch2015}. Prior works argue that adaptivity in such scenarios depends on perception to localize materials, compare as-built and as-planned states, and correct actions based on measurements~\cite{adel2024}. Existing robotic systems for aiding masonry construction, such as Hadrian~\cite{noauthor_hadrian_nodate}, leverage specialized machinery and standardized materials to circumvent this issue. For robot-aided masonry construction, the key challenge remains capturing and correcting geometric tolerances in situ before they propagate. Building on feedback-driven adaptive construction frameworks~\cite{Ruan2023, adel2024}, this paper proposes an adaptive pipeline for human--robot collaborative masonry construction in which laser sensing is used to measure brick poses, brick dimensions, and as-built course heights, thereby enabling precise grasping and placement while maintaining level courses and aligned wall edges.

\begin{figure*}[!htb]
    \centering
    \includegraphics[width=1\textwidth]{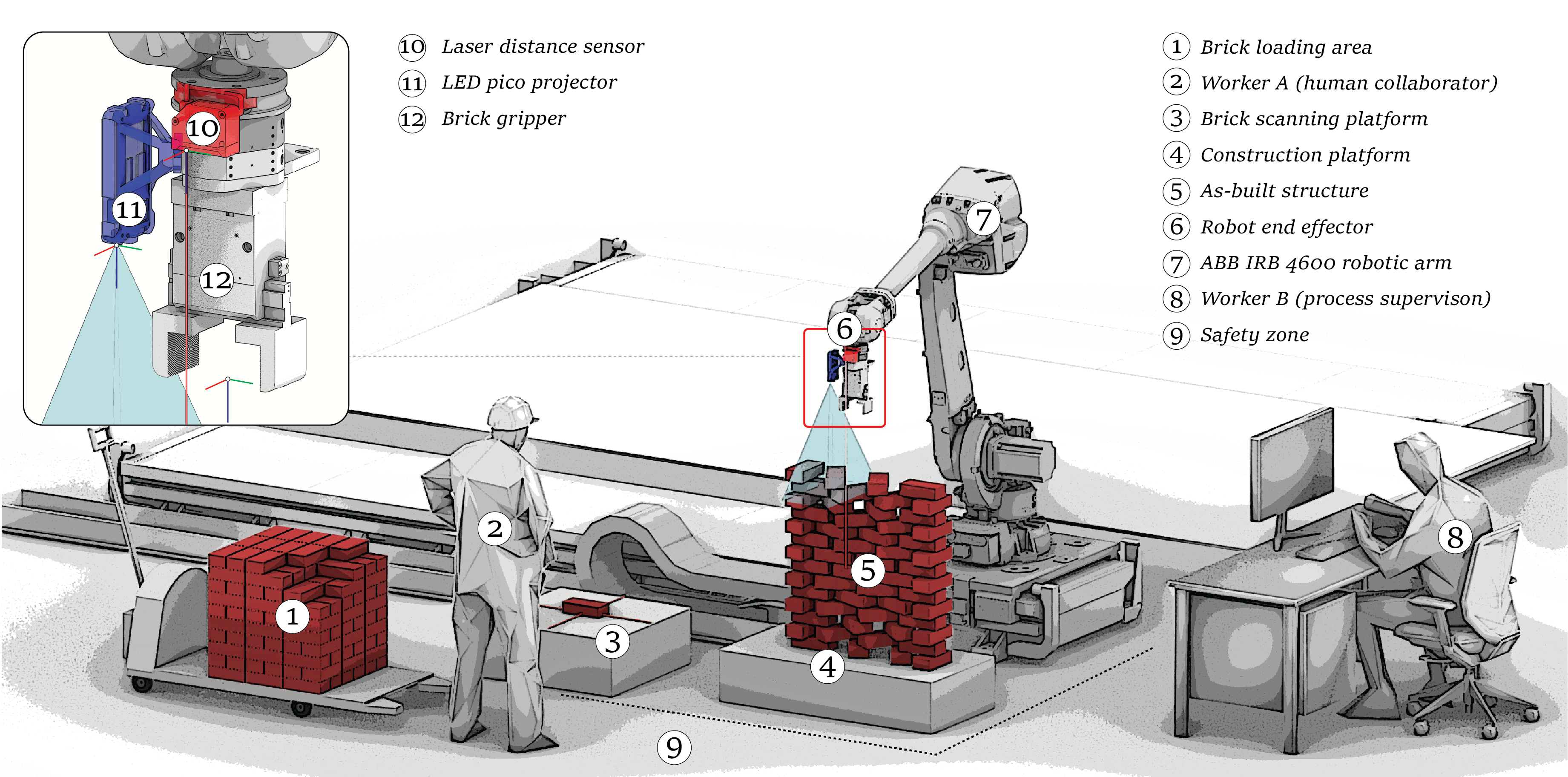}
    \caption{Workcell setup for augmented human--robot collaborative masonry construction.}
    \label{fig:system_overview}
\end{figure*}

\section{Methods}
\label{sec:method}

The methods section describes the implementation of a human--robot collaborative masonry workflow and is organized to detail three components of the system: the workcell setup (Section~\ref{sec:system_overview}); the modeling, mapping, and calibration of the end-effector--mounted projector for delivering spatially explicit visual cues (Section~\ref{sec:projection}); and the adaptive pipeline (Section~\ref{sec:adaptive_construction}). Collectively, these components enable human guidance and robot adaptation, improving robustness and construction accuracy. 

\subsection{Workcell Setup}
\label{sec:system_overview}

\begin{figure}[!htb]
    \centering
    \includegraphics[width=0.5\textwidth]{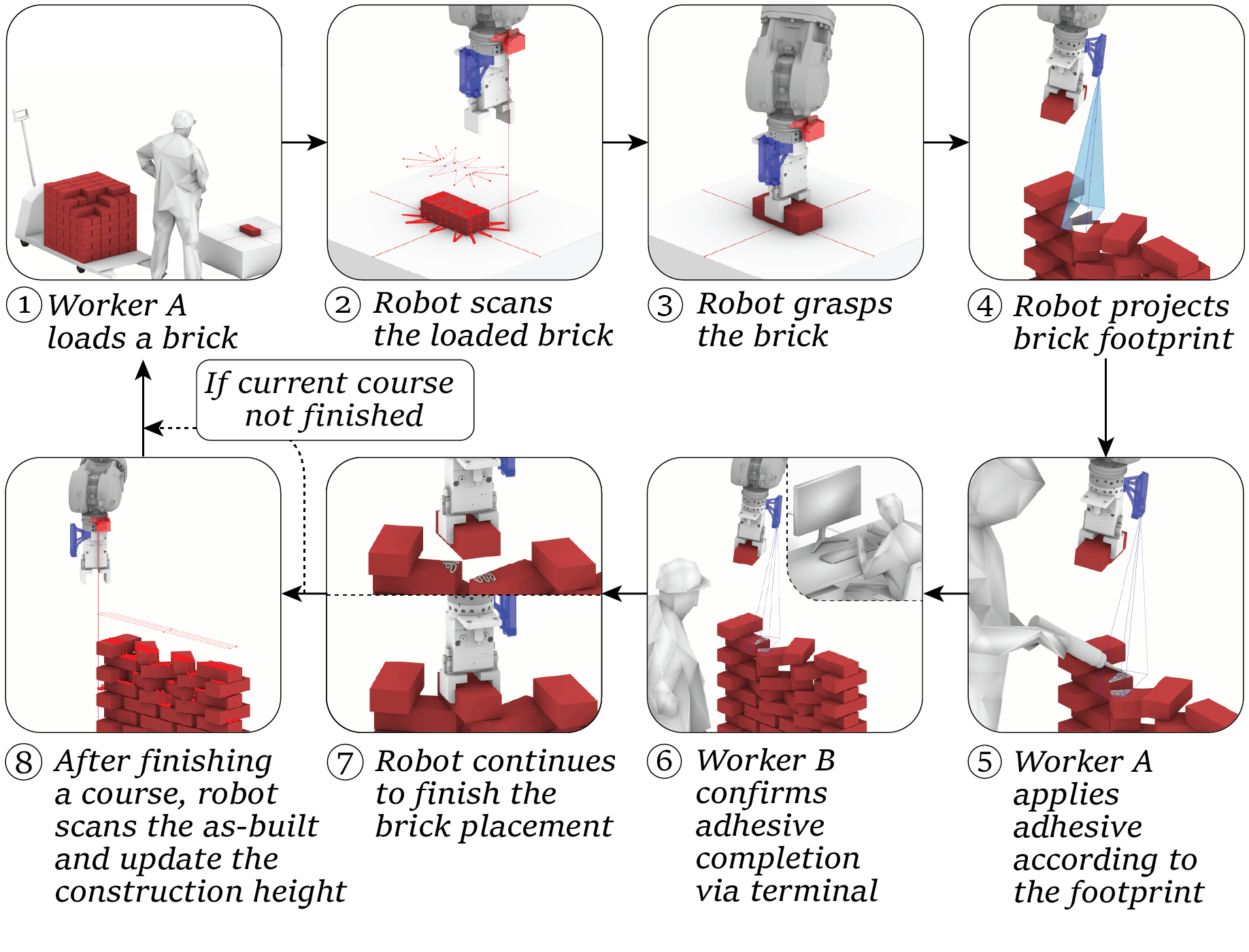}
    \caption{Collaborative masonry workflow integrating projection for guiding human action.}
    \vspace{-20pt}
    \label{fig:workflow}
\end{figure}

The robotic system operates in a shared human--robot workcell comprising a brick loading area, pickup platform, construction platform, and designated human safety zone (Figure~\ref{fig:system_overview}). Two workers participate in the process: Worker~A applies adhesive and performs local safety checks, while Worker~B monitors system state, conducts global safety checks, and coordinates the robot. The construction process follows a repetitive pick-and-place cycle that combines robotic precision with human dexterity through spatially explicit communication (Figure~\ref{fig:workflow}). Before grasping each brick, the robot scans the brick's top surface with a one-dimensional laser sensor to estimate its pose and dimensions. After grasping, the robot then moves to a projection pose and displays the planned brick footprint as a registered quadrilateral using the end-effector--mounted projector (Section~\ref{sec:projection}). Worker~A applies adhesive within the projected boundary, confirms completion with Worker~B, and the robot resumes placement once the worker returns to the safety zone. Between courses, the robot scans the as-built surface with the laser sensor and applies corrections derived from these measurements to subsequent placements (Section~\ref{sec:adaptive_construction}).

\subsection{Projector Integration for Spatially Explicit Visual Cues in Human--Robot Collaboration}
\label{sec:projection}

Prior work has integrated projector--camera units with robotic mechanisms, using projector--camera calibration, camera-based pose estimation, and kinematic control to register projected information onto physical surfaces~\cite{lee_data-driven_2016}. Rather than estimating projector's relative pose using camera observations, we estimate the projector--end-effector transformation through robotic contact probing and robot-kinematics--based pose estimation. Conceptually, this approach draws on hand--eye calibration~\cite{tsaiNewTechniqueFully1989}, in which the fixed transformation between an eye-on-hand camera and the robot gripper is recovered from camera observations across multiple robot poses. We adapt this calibration logic to estimate the transformation between the gripper-mounted projector and the robot end-effector, enabling just-in-time, spatially registered visual cue display for human--robot collaborative assembly.

We define each spatial visual cue as a set of 3D points $\mathcal{Q}=\{\mathbf{q}_{j}\in\mathbb{R}^3\}$ with $j = 1,2,3...$ in the world frame that encodes task-relevant information, e.g., locations of fasteners, centroids of surfaces, and, in the case of this study, a brick footprint for showing the placement plan. Let $\{W\}$ denote the world frame, $\{T\}$ the gripper tool center point (TCP) frame, and $\{P\}$ the projector optical frame, whose origin coincides with the projector’s effective viewing cone's apex at the lens and whose Z axis aligns with the direction of projection. The projector is rigidly mounted to the side of the gripper, and its pose with respect to the gripper's TCP is represented by a fixed homogeneous transformation ${}^{T}\mathbf{T}_{P} \in SE(3)$, which is obtained from the projector calibration process. The desired and reachable projector pose ${}^{W}\mathbf{T}_{P}$ is generated given $\mathcal{Q}$ and the robot base pose to ensure full projection coverage and a proper projection distance, which, in this study, is achieved through a fixed translation of $\mathcal{Q}$'s bounding box centroid for pose origin and directing the pose Z axis to the centroid. In operation, given ${}^{T}\mathbf{T}_{P}$ and ${}^{W}\mathbf{T}_{P}$, we calculate the target TCP pose ${}^{W}\mathbf{T}_{T}$ to position the projector: 

\begin{equation}
{}^{W}\mathbf{T}_{T} = {}^{W}\mathbf{T}_{P}\,{}^{T}\mathbf{T}_{P}^{-1}
\end{equation}

\begin{figure}[!htb]
    \centering
    \includegraphics[width=0.5\textwidth]{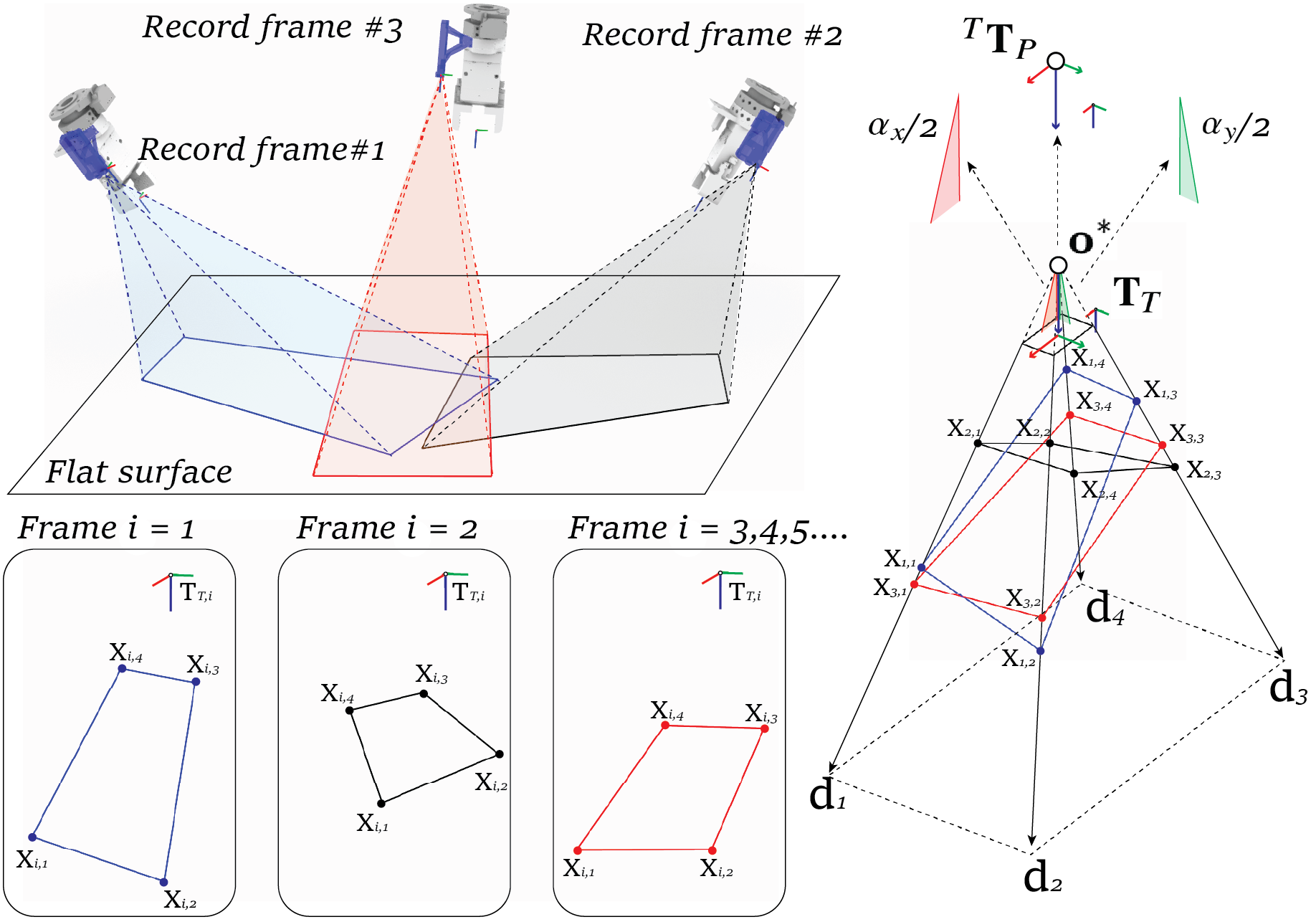}
    \caption{Projector calibration: finding the accurate transformation from the gripper TCP pose to the projector pose.}
    \vspace{-20pt}
    \label{fig:calibration}
\end{figure}

To properly display the spatial visual cue $\mathcal{Q}$ through projection, the projector is modeled as a reversed pinhole device, analogous to an inverse camera~\cite{tsai_versatile_1987, zhang_projection_2006}, with its optical center treated as a point source due to its small aperture relative to the projection distance. The intrinsic parameters of the projector include horizontal and vertical field-of-view angles (FOV), $(\alpha_x,\alpha_y)$, as shown in Figure~\ref{fig:calibration}. For a point in $\mathcal{Q}$ expressed in the projector frame,  ${}^{P}\mathbf{q} = [X, Y, Z]^\top$, the corresponding normalized 2D coordinates on the projection image plane, $(u,v)\in[0,1]^2$ are computed via:

\begin{equation}
u = \frac{1}{2}\left(1 + \frac{X}{Z\tan(\frac{1}{2}\alpha_x)}\right), \quad
v = \frac{1}{2}\left(1 + \frac{Y}{Z\tan(\frac{1}{2}\alpha_y)}\right)
\end{equation}

\noindent where $u=0, 1$ correspond to the image plane's left and right bounds, and $v=0, 1$ correspond to the image plane's top and bottom bounds. 

To obtain accurate ${}^{T}\mathbf{T}_{P}$ and the FOV $(\alpha_x,\alpha_y)$, calibration is performed using a sufficiently large planar surface and a contact probe rigidly attached to the end-effector for locating its TCP physically. The process is shown in Figure~\ref{fig:calibration}. A full-frame rectangle is cast onto the planar surface from more than two projection poses, ensuring that all four projected corners lie on the surface. For projection $i = 1, 2, 3...$, the robot TCP pose ${}^{W}\mathbf{T}_{T,i}$ is recorded, and the four projected corner locations are then marked on the surface and measured using the contact probe, yielding 3D coordinates ${}^{W}\mathbf{X}_{i,k}$ in the world frame, where $k \in \{1,2,3,4\}$. All measured points are transformed into the TCP frame via homogeneous transformation:

\begin{equation}
{}^{T}\mathbf{X}_{i,k} =
{}^{W}\mathbf{T}_{T,i}^{-1}\;
{}^{W}\mathbf{X}_{i,k}
\end{equation}

For each corner $k$, a line in 3D space is fitted to the corresponding point set $\{{}^{T}\mathbf{X}_{i,k}\}$ using principal component analysis, yielding four lines that approximate the edges of the projector's viewing cone in the TCP frame (Figure.~\ref{fig:calibration}). The projector optical center is estimated as the point $\mathbf{o}^\ast$ that minimizes the sum of squared distances to these lines:

\begin{equation}
\mathbf{o}^\ast =
\arg\min_{\mathbf{o}}
\sum_{k=1}^{4}
\left\|
\left(\mathbf{I} - \mathbf{d}_k\mathbf{d}_k^\top\right)
(\mathbf{o} - {\mathbf{x}}_k)
\right\|^2
\end{equation}

\noindent where $\mathbf{x}_k$ is a point on the $k$-th line and $\mathbf{d}_k$ is its unit direction vector. The projector optical axis is computed as the normalized average direction of the fitted lines. The projector's principal image axes are reconstructed by averaging the directions of opposing line pairs. The two principal image axes are then orthonormalized with the optical axis to form the projector pose ${}^{T}\mathbf{T}_{P}$ in the TCP frame. The projection FOV $(\alpha_x,\alpha_y)$ is obtained geometrically from the reconstructed viewing cone.

\subsection{Adaptive Construction Pipeline}
\label{sec:adaptive_construction}

The adaptive approach here is formulated as pose correction enabled through perception. The pipeline operates at two levels, as presented in Figure~\ref{fig:perception}: brick-level adaptation to estimate individual brick pose and dimensions before grasping; and course–level adaptation, reflected in brick-placement pose correction to compensate for cumulative tolerances across courses. In this section, let $\{W\}$ denote the world frame, $\{b\}$ the brick's pose, $\{g\}$ the gripper TCP pose when grasping and $\{r\}$ when placing and releasing the brick. 

\subsubsection{Brick-Level Pose Estimation and Adaptive Grasping}
\label{sec:brickLevel}
For each brick, a predefined scan path is used to acquire a point cloud measurement of the brick's top face. The point cloud is then fitted with a minimum-area bounding box to estimate the brick’s pose and dimensions. Let ${}^{W}\mathbf{T'}_{b} \in SE(3)$ denote the updated brick pose through scanning. This pose defines the grasping frame via:

\begin{equation}
{}^{W}\mathbf{T'}_{g} =
{}^{W}\mathbf{T'}_{b} \, {}^{b}\mathbf{T}_{g}
\end{equation}

\noindent where predefined ${}^{b}\mathbf{T}_{g}$ encodes a fixed TCP grasping pose in the brick's frame.

\subsubsection{Course-Level Alignment and Placement Correction}

Before placing each course of bricks, the robot scans the top of the current course to estimate the as-built height by averaging the heights of the scanned points. Nominal placement poses ${}^{W}{\mathbf{T}}_{r}$ are precomputed from the design model, including reference vertical planes defining wall-edge alignments. Vertical adaptation adjusts the placement height based on the measured course top height, the scanned brick height (measured in Section~\ref{sec:brickLevel}), and a constant adhesive thickness. Let $z'$ be the measured surface height, $h'_{b}$ the scanned brick's height, and $h_{\mathrm{a}}$ the predefined adhesive thickness (0.8 mm). The adjusted placing height is:

\begin{equation}
z'_{r} =
z' +
h'_{b} +
h_{\mathrm{a}}
\end{equation}

\begin{figure}[!htb]
    \centering
    \includegraphics[width=0.5\textwidth]{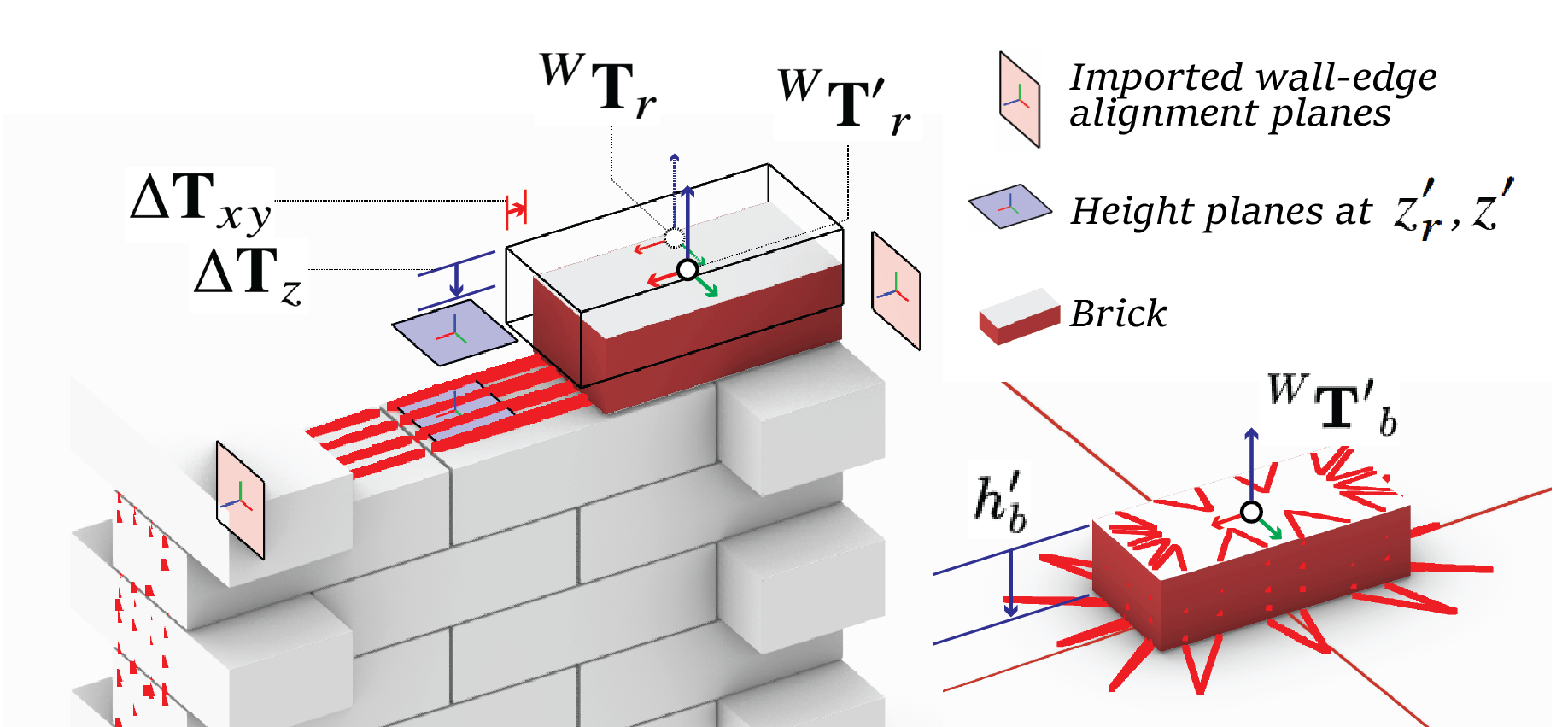}
        \caption{Brick pose correction based on brick-level and course-level scanning. Left: Placing correction. Right: Grasping correction.}
    \vspace{-20pt}
    \label{fig:perception}
\end{figure}

Horizontal adaptation enforces wall-edge alignments by adjusting in-plane placement to compensate for perceived brick length variations (Section~\ref{sec:brickLevel}). To accommodate this correction, a 3-mm lateral gap is introduced between the nominal brick geometries. The final corrected placement pose for each brick is:

\begin{equation}
{}^{W}\mathbf{T'}_{r} =
{}^{W}{\mathbf{T}}_{r} \, \Delta \mathbf{T}_{z} \, \Delta \mathbf{T}_{xy}
\label{eq:adaptive_pose}
\end{equation}

where $\Delta \mathbf{T}_{z}$ is a vertical correction shared within a course and $\Delta \mathbf{T}_{xy}$ is a brick-specific lateral correction. Defining brick poses at the brick top center and apply vertical corrections enforces level courses, while lateral corrections maintaining wall-edge alignment under bricks' dimensional variability (Figure~\ref{fig:perception}).

\section{Experimental Setup}\label{sec:experiments}

\begin{figure}[!htb]
    \centering
    \includegraphics[width=0.3\textwidth]{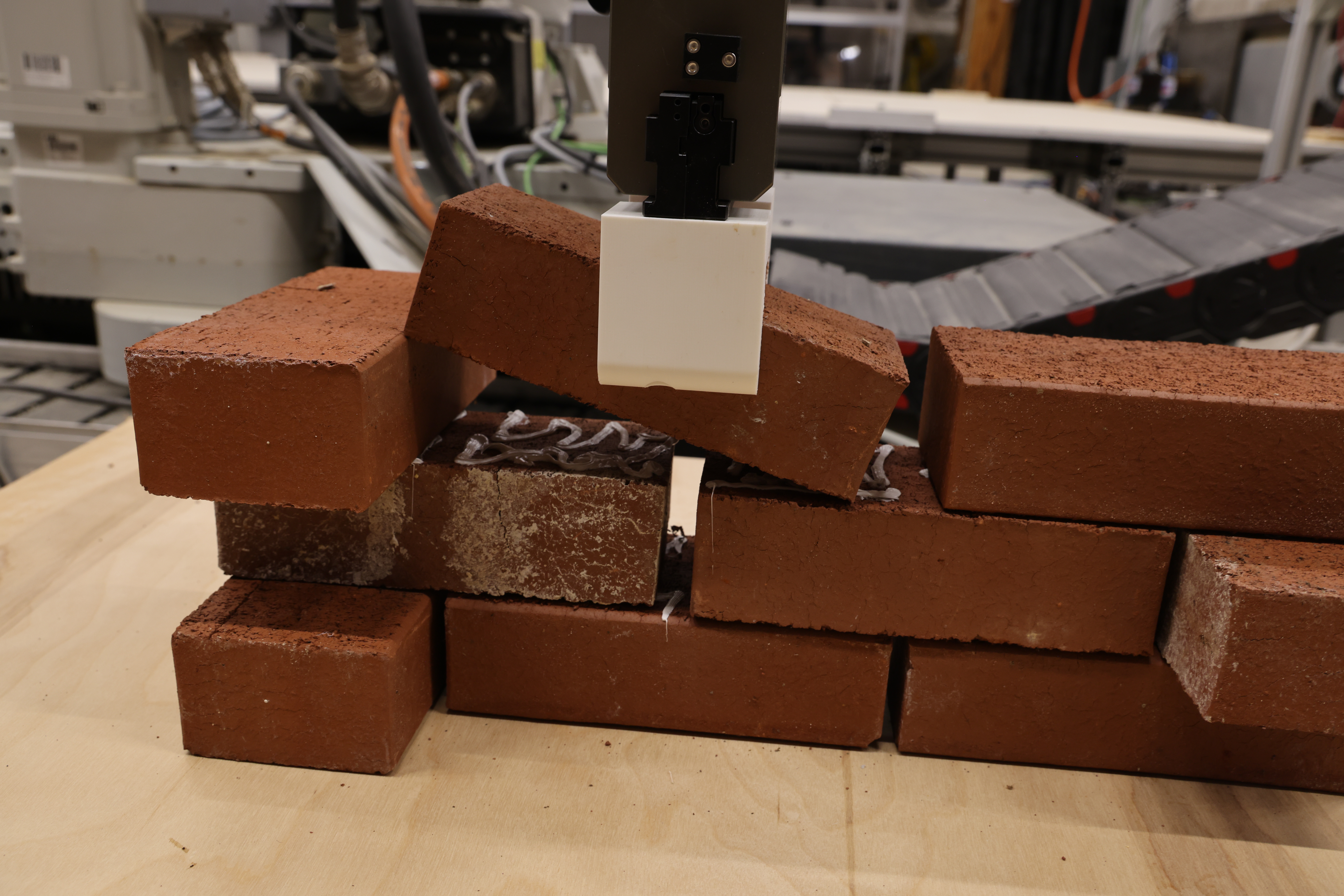}
    \caption{The experiment without brick pose correction leads to collision and early failure.}
    \vspace{-20pt}
    \label{fig:failure}
\end{figure}

\begin{figure}[!htb]
    \centering
    \includegraphics[width=0.45\textwidth]{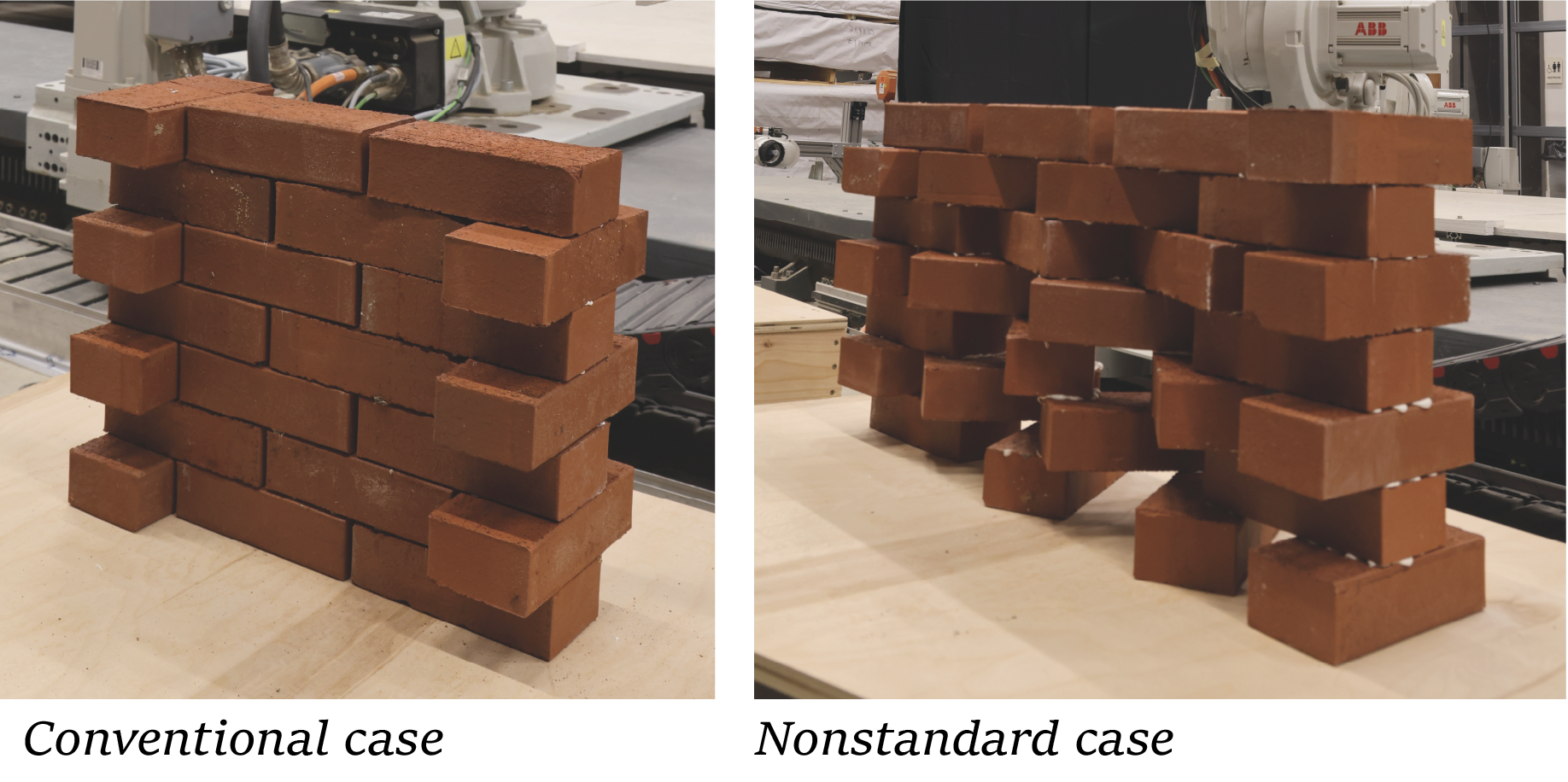}
    \caption{Variations in brick configuration, including conventional (straight running bond) and nonstandard (running bond with brick rotations).}
    \vspace{-20pt}
    \label{fig:configs}
\end{figure}

\begin{figure*}[!htb]
    \centering
    \includegraphics[width=1\textwidth]{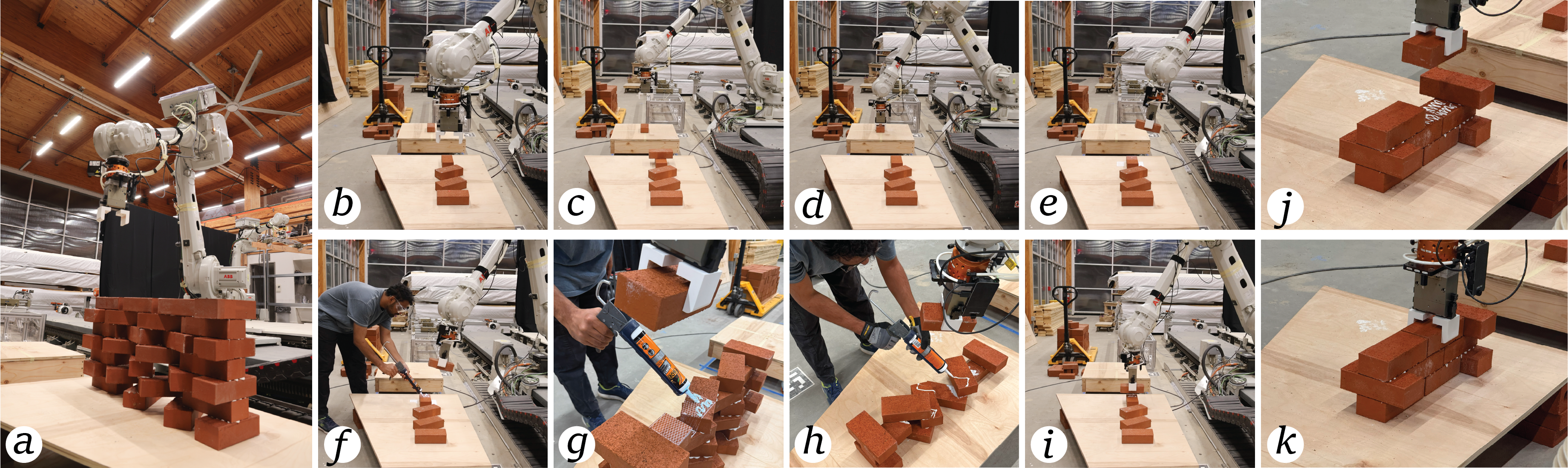}
    \caption{Documented experiment processes: a. robotic manipulator and a finished nonstandard construct; b. course scanning; c. brick scanning; d. brick grasping; e. footprint projection; f. adhesive application; g. projection-augmented condition; h. unaugmented condition (human collaborator trying to predict brick pose); i. brick placement; j \& k: brick placement in the tightly constrained conventional configuration.}
    \vspace{-20pt}
    \label{fig:process_photos}
\end{figure*}

An initial experiment without brick pose correction failed due to a collision resulting from accumulated grasping and placement tolerances (Figure~\ref{fig:failure}), underscoring the necessity of the adaptive pipeline for collaborative masonry. Four successful experiments, integrated with brick pose correction, were then conducted, varying brickwork configurations and projection usage. Two configurations were tested: a nonstandard assembly of 28 bricks over seven courses and a conventional running-bond assembly of 21 bricks over seven courses, as presented in Figure~\ref{fig:configs}. For each, two conditions were evaluated: \emph{projection-augmented}, with the footprint projected during adhesive application, and \emph{projection-verified}, with projection used only for verifying adhesive application quality. The experiment processes are documented in Figure~\ref{fig:process_photos}. 

\subsection{Evaluation Metrics}

The effectiveness of projection augmentation in supporting manual adhesive application was first evaluated. For each brick, images of the adhesive patch with the projected footprint were analyzed using a vision pipeline to compute: 1) region coverage (\%), the fraction of the intended bonding area covered by identified adhesive pixels; 2) exposed adhesive (\%), the fraction of identified adhesive pixels applied outside the target region; and 3) largest missed patch (\%), defined as the largest contiguous uncovered area ratio within the target bonding region. These metrics jointly capture adhesive completeness, cleanliness, and continuity. The assemblies' geometric accuracy was evaluated from course-level laser scans. The scanned point clouds were segmented into brick faces, and the mean absolute error of point heights relative to the updated target heights was computed, reflecting the effectiveness of brick pose correction in enforcing level courses. Process efficiency was assessed by logging timestamps for each pick-and-place cycle and sub-processes, including robot traversal, laser scanning, data collection, adhesive application, and user-input wait time. This breakdown identifies time-consuming stages and quantifies the effects of projection guidance and adaptive sensing on total task duration.

\begin{table*}[t]
\centering
\caption{Adhesive Coverage Analysis}
\label{tab:coverage}
\renewcommand{\arraystretch}{1.9}
\setlength{\tabcolsep}{4pt}
\begin{tabular}{L{5.2cm} L{2.4cm} L{2.4cm} L{2.4cm} L{2.4cm}}
\hline
\textbf{Experiment condition} &
\textbf{Region coverage(\%)} &
\textbf{Exposed adhesive(\%)} &
\textbf{Largest missed patch(\%)} &
\textbf{Application time (s)} \\
\hline
Nonstandard, Projection Augmented & 49.1 ± 9.1 & 2.4 ± 2.1 & 7.2 ± 13.3 & 25.2 ± 11.7 \\
Nonstandard, Projection Verified & 39.9 ± 12.6 & 19.6 ± 22.0 & 17.8 ± 23.2 & 26.9 ± 9.0 \\
Conventional, Projection Augmented & 40.5 ± 5.2 & 0.6 ± 0.8 & 10.2 ± 6.4 & 15.1 ± 8.4 \\
Conventional, Projection Verified & 40.1 ± 4.2 & 0.7 ± 0.6 & 9.4 ± 5.1 & 24.7 ± 9.0 \\
\hline
\end{tabular}
\end{table*}

\begin{figure}[!htb]
    \centering
    \includegraphics[width=0.5\textwidth]{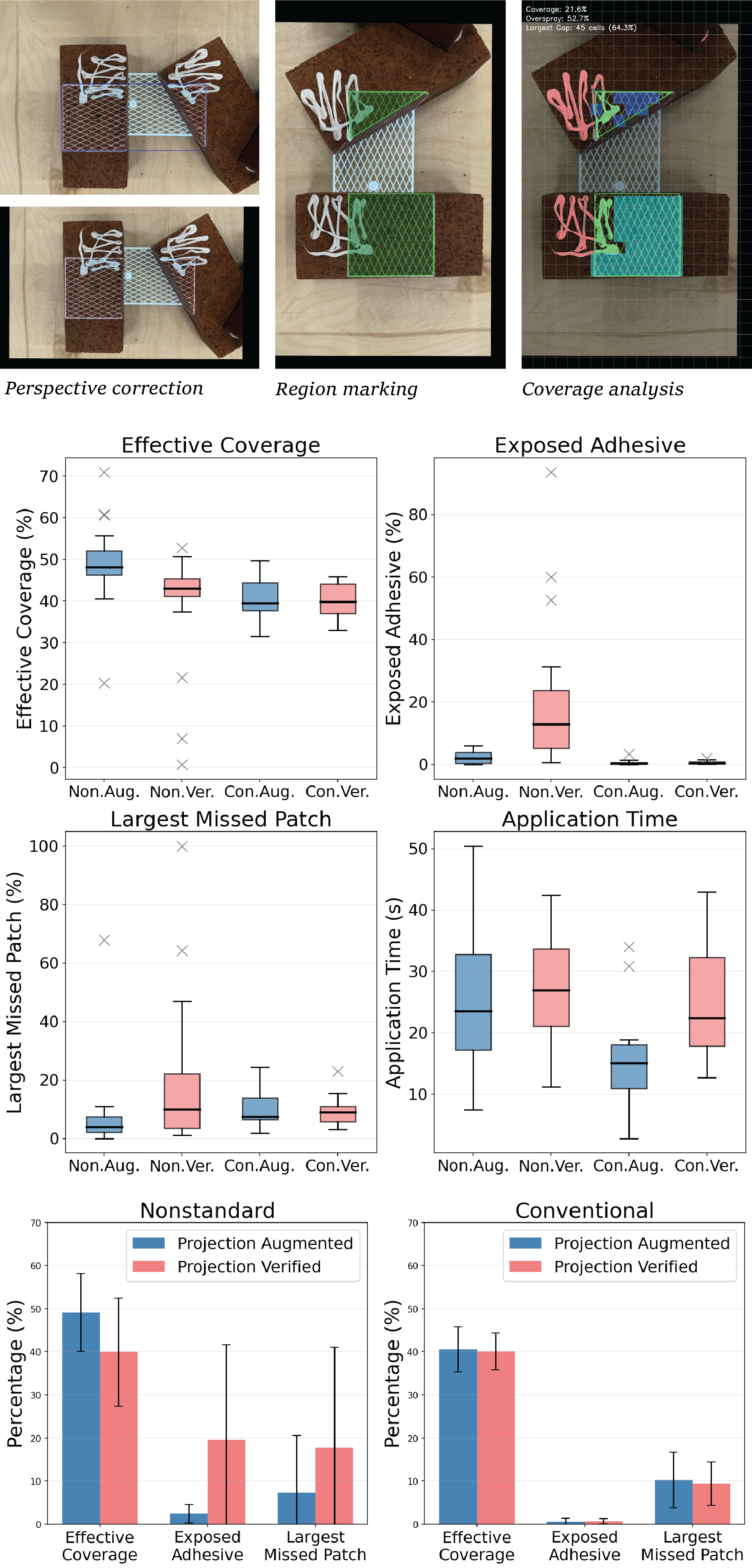}
    \caption{Adhesive coverage identification and analysis. Numerical values are reported in Table~\ref{tab:coverage}.}
    \vspace{-20pt}
    \label{fig:coverage}
\end{figure}

\label{sec:accuracy}
\begin{figure}[!htb]
    \centering
    \includegraphics[width=0.44\textwidth]{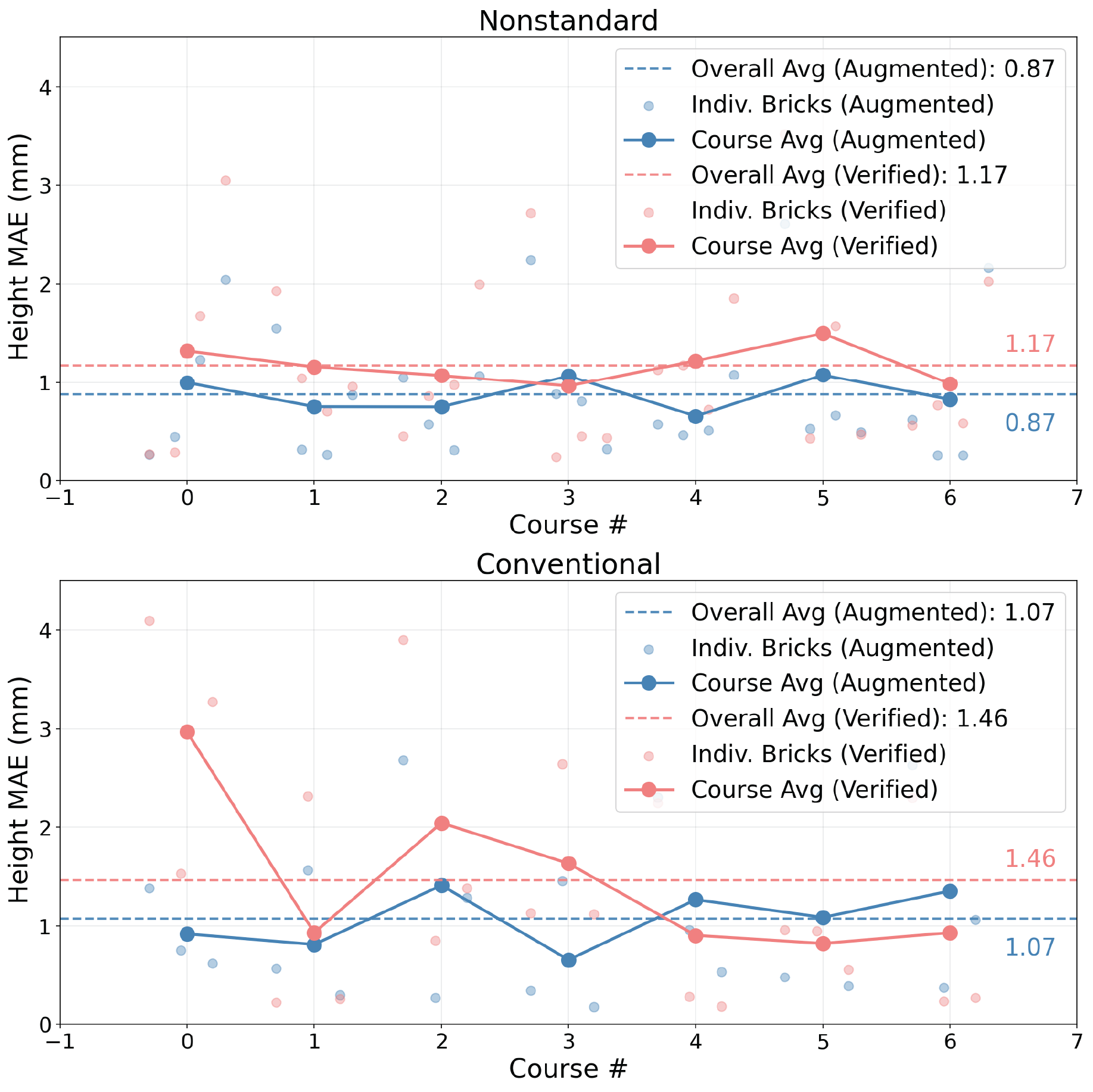}
    \caption{Height mean absolute error (MAE) analysis based on course-level scanning results.}
    \vspace{-20pt}
    \label{fig:accuracy}
\end{figure}

\section{Results and Discussion}
\label{sec:results}

Quantitative results on adhesive quality and application time are summarized in Table~\ref{tab:coverage} and Figure~\ref{fig:coverage}. Across the experiments, effective adhesive coverage generally stabilized between 40--50\%. However, the nonstandard projection-verified case showed large fluctuations, including some near-zero values. Together with the higher coverage observed in the nonstandard projection-augmented case, these results suggest that projection guidance improves the consistency of human-applied adhesive coverage in the nonstandard case. Similar trends are observed in exposed adhesive and the largest missed patch metrics. Compared with the projection-augmented case, the projection-verified nonstandard case exhibits highly scattered exposed-adhesive ratios (up to $>90\%$) and higher missed-patch ratios, indicating imprecise application and material waste. These findings suggest that projection augmentation improves the quality of human-applied adhesive for nonstandard brickwork. For the conventional running-bond configuration, adhesive quality was comparable with and without projection. However, the reduced application time in the augmented case indicates faster execution without loss of quality. Overall, the method is shown to effectively augment human performance by improving adhesive application accuracy in nonstandard brickwork and application speed in conventional brickwork.

As shown in Figure~\ref{fig:accuracy}, the brick pose correction process limited height deviations to approximately 1.0 mm. This demonstrates effective cancellation of cumulative assembly tolerances. With brick pose correction, collision failures due to inaccurate placement (Figure~\ref{fig:failure}) were not observed, indicating improved placement quality. Projection-augmented nonstandard cases achieved a lower average mean absolute error than projection-verified cases, suggesting the significance of precise adhesive application in reducing course height deviations.

The time analysis shows that, across all experiments, the average time to process a single brick is 123.1~s. Aside from data collection (14.2\%) and adhesive application (15.9\%), robot traversal accounts for over 42.6\% of the total cycle time, largely due to the conservative motion speed (250~mm/s) adopted for safety. User input waiting was the second-largest contributor (20.3\%), revealing limitations of terminal-based interaction and motivating future multimodal communication. At this preliminary stage, a 1D laser sensor was used for ease of implementation; data collection and scanning overheads are expected to decrease (currently 7.0\%) with higher-dimensional sensing, as shown in prior work~\cite{adel2024}.

\section{Conclusion}
\label{sec:conclusion}

This work presents a human--robot collaborative masonry workflow that integrates projection-based robot-to-human guidance with feedback-driven adaptive brick placement. Using adhesive-based brickwork as the case study, an end-effector-mounted projector displays desired adhesive regions directly on the work-in-progress structure, while laser-based sensing supports brick-level grasp correction and course-level placement adaptation. Comparative experiments show that projection guidance improves adhesive application accuracy in nonstandard configurations and reduces application time in conventional running-bond patterns. At the same time, the adaptive pipeline maintains course-height accuracy and avoids collision-prone failures associated with open-loop execution. Together, these results demonstrate how projection-augmented interaction and feedback-driven adaptation can improve precision and robustness in human--robot collaborative masonry construction.

Current system includes limited integration of active safety monitoring and a constrained human-to-robot communication interface, which currently relies on indirect terminal-based input. Future work will expand the system in several directions: incorporating direct communication modalities, such as gesture and speech, to support more intuitive collaboration; integrating 2D/3D perception to reduce scanning overhead; and adding perception-based safety modules for more robust operation. In addition, a user study with worker feedback is needed to evaluate system performance from a human-centered perspective. 

\section*{Acknowledgments}

This research was supported by Princeton University. The work presented in this paper builds on a course project developed in collaboration with Yilin Lin and Samet Yilmaz. We thank Daniel Ruan and Salma Mozaffari for their feedback and technical support, and the members of the Adel Research Group for their assistance during system development and experiments.

\bibliography{main}

\end{document}